\documentclass[10pt,twocolumn,letterpaper]{article}

\usepackage{cvpr}              
\definecolor{cvprblue}{rgb}{0.21,0.49,0.74}
\usepackage[pagebackref,breaklinks,colorlinks,allcolors=cvprblue]{hyperref}

\def\paperID{} 
\def\confName{CVPR}
\def\confYear{2026}

\title{SketchAssist: A Practical Assistant for Semantic Edits and Precise Local Redrawing}

\author{
Han Zou\qquad
Yan Zhang${^*}$${^{\dagger}}$\qquad
Ruiqi Yu\qquad
Cong Xie\qquad
Jie Huang\qquad
Zhenpeng Zhan${^{\dagger}}$\qquad\\[0.5ex]
Global Business Unit, Baidu Inc.
\\ 
\texttt{\small \{zouhan, zhangyan97, yuruiqi, xiecong, huangjie13, zhanzhenpeng01\}@baidu.com} \\
}
\newcommand\extrafootertext[2]{%
    \bgroup
    \renewcommand\thefootnote{\fnsymbol{footnote}}%
    \renewcommand\thempfootnote{\fnsymbol{mpfootnote}}%
    \footnotetext[#1]{#2}%
    \egroup
}

\begin{document}
\maketitle

\extrafootertext{1}{Project leader. $^{\dagger}$Corresponding authors.}

\begin{abstract}
Sketch editing requires jointly handling high-level semantic changes and precise local redrawing, a combination that is particularly challenging for sparse, style-sensitive line art. Unlike natural images, sketches rely on minimal visual cues, making it difficult for existing methods to reconcile global semantic modifications with fine-grained structural control while preserving overall coherence.  We present \textbf{SketchAssist}, an interactive sketch assistant that unifies instruction-guided editing with line-guided region redrawing, enabling efficient and controllable sketch manipulation while preserving overall composition.

To support this task, we introduce a controllable data generation pipeline that constructs structured edit sequences with precise attribute variations and maintains structural alignment across multi-step modifications, while expanding stylistic diversity via style-preserving transformations.

Building on this data, SketchAssist adopts a unified framework based on DiT~\cite{Dit}, using a multi-channel input representation to encode sketches, masks, and guidance signals within a single interface. To further handle different editing modes, we integrate a Task-guided Mixture-of-Experts (T-MoE) into LoRA layers, enabling adaptive control over semantic and structural guidance.

Extensive experiments demonstrate state-of-the-art performance on both tasks, achieving strong instruction adherence and improved structural and style consistency compared to recent methods. Together, our method provide a practical and controllable solution for sketch editing.
\end{abstract}    
\section{Introduction}
\label{sec:intro}

Sketch editing plays a fundamental role in digital art creation. In professional practice, sketch creation is structured as a two-phase, layout-to-detail process. \textit{Phase 1 (layout):} the artist establishes global composition—pose, body proportions, and framing—iteratively revising until the intended narrative is fixed. Fine local attributes (e.g., eyes, hair, background textures) are deliberately deferred; adding them before the layout stabilizes hinders iteration and wastes effort. \textit{Phase 2 (detail):} with the composition fixed, the artist incrementally adds and edits local visual attributes, refining line art and secondary structure while preserving the established pose and composition.

The second stage is where artists spend most strokes and time. To improve efficiency, we harness modern image-generation models to automate the second stage via two complementary interfaces:

\noindent\textbf{Instruction-Guided Sketch Editing} —
Given a source sketch and a natural-language instruction, the goal is to add, remove, or replace visual attributes of the character or scene (e.g., outfit, hairstyle, background) while keeping unrelated regions and the overall composition unchanged. This interface emphasizes semantic controllability: artists describe \emph{what} should change, and the model proposes a plausible edit that respects the original sketch.

\noindent\textbf{Line-Guided Region Redrawing} —
For edits that are \emph{not easily expressible in text} alone, the user additionally provides a binary mask and guidance lines. These guidance strokes (or reference line art) encode explicit spatial and structural cues such as contours and proportions, enabling precise synthesis that remains coherent with the surrounding (out-of-mask) line art and overall style.

Although existing image editing and inpainting models have the potential to perform the tasks described above, they still fall short in meeting the specific requirements of sketch editing, particularly in jointly handling semantic and structural guidance, with challenges in both data and model aspects. From a data perspective, most current editing datasets are derived from real images and do not generalize well to the sparse, line-based nature of sketches. More importantly, existing sketch datasets rarely provide controllable paired samples that depict the same subject under different states with precise attribute changes. This is particularly true for complex, multi-attribute modifications, where simple synthetic strategies fail to maintain structural alignment across sequential edits. This gap motivates our controllable data generation pipeline.

From a model perspective, instruction-based editing typically influences the entire image, whereas localized redrawing demands precise control over targeted regions. Bridging these two modes within a unified framework—while maintaining style fidelity across diverse artistic styles—remains an open challenge, as the model must balance high-level semantic instructions with low-level structural constraints.

To address these challenges, we first design a controllable data generation pipeline tailored for sketches. We establish a foundation of reliability by constructing atomic addition sequences, which are then expanded into complex editing scenarios via a cross-sequence sampling strategy. This approach allows us to synthesize supervised pairs spanning from single-attribute changes to multi-attribute transformations, effectively covering add, remove, and replace operations within a unified edit distance framework. To further ensure stylistic robustness, we complement these synthetic sequences with style-diversified pairs derived from diverse real-world sources using a style-preserving attribute modification strategy. For line-guided redrawing, we leverage extracted linework as structural hints, teaching the model to harmonize user-provided strokes with the original artistic style.

Building upon this data foundation, we propose SketchAssist, a unified framework designed to handle both instruction-guided editing and line-guided region redrawing within a single architecture. SketchAssist adopts a novel unified conditional representation, implemented as RGB channels, that encodes the source sketch, that encodes the source sketch, target mask, and user strokes into a structured input, enabling the network to seamlessly interpret heterogeneous guidance signals. To accommodate the distinct patterns of the two tasks and mitigate task interference, we introduce a Task-guided Mixture-of-Experts (T-MoE) module that allows the model to dynamically adapt its parameters to the corresponding editing mode. This design achieves state-of-the-art performance on both tasks, offering artists a flexible and stylistically coherent creative assistant.

\textbf{Our main contributions are:}
\begin{enumerate}
\item We develop a sketch-specific data generation pipeline that produces controllable editing pairs via atomic attribute sequences and cross-sequence sampling, enabling diverse and structurally aligned edits.
\item We introduce a unified input representation that encodes source sketches, masks, and guidance within a shared multi-channel format, enabling seamless switching between editing modes.
\item We propose SketchAssist, a unified framework with a task-guided Mixture-of-Experts (T-MoE) architecture that handles both semantic and structural guidance for precise and controllable sketch editing.
\end{enumerate}

\section{Related Works}
\label{sec:related_wokrks}

Existing sketch editing studies can be broadly categorized into two directions: \textbf{instruction-based image editing} and \textbf{local image editing/redrawing}, corresponding to the two complementary modes in our target task—global edits guided by textual or semantic instructions, and localized repaints guided by user-provided strokes.

\textbf{Instruction-based image editing} modifies images according to natural language prompts, enabling global semantic manipulation such as changing character identity, actions, or background. Early approaches combined GANs~\cite{Generative,Large,Alias-free,A-style-based-generator,Analyzing} with CLIP~\cite{Clip} for semantic guidance, followed by VQ-GAN~\cite{Taming,Vqgan-clip} for more diverse generation. Recent works leverage diffusion models~\cite{Denoising,SD,hertz2022prompt,kawar2023imagic,brooks2023instructpix2pix,Sheynin_2024_CVPR,jintp}, particularly those based on large-scale transformers such as DiT~\cite{Dit,FLUX,Qwen,Enabling,liu2025step1x} to achieve state-of-the-art quality and flexibility. However, these methods remain limited in fine-grained local control.

\textbf{Local image editing and redrawing} preserves overall content while selectively modifying specified regions, maintaining stylistic and structural coherence. Mask-based methods~\cite{Blended,Repaint,Brushnet} are straightforward but lack intuitive direct-editing flexibility. Alternatives include text-based localized editing~\cite{Text2live,Region-Prompt}, sketch-based control~\cite{Sketchedit}, and multimodal inputs combining masks, text, and edges~\cite{Magicquill}. These approaches offer better local control but often lack the global semantic manipulation power of instruction-based frameworks. Building on an instruction-based backbone, our work integrates line-guided local editing to provide precise regional control while retaining strong global semantic manipulation capabilities.

\section{Data Generation}
\label{sec:data}

\subsection{Problem Definition}
Here, we formalize the two complementary tasks of our system, specifying their inputs, outputs, and core objectives.

\noindent\textbf{Instruction-Guided Sketch Editing} —  
Given a source sketch image $I_s \in \mathbb{R}^{H \times W \times 3}$ and a textual instruction $t$, the goal is to precisely add, remove, or replace specific visual attributes of the depicted character (e.g., hairstyle, clothing, facial expression) according to $t$. This task emphasizes \textit{semantic controllability}: the model must correctly interpret instructions of varying complexity (from single-attribute changes to multi-attribute combinations) while strictly preserving unrelated regions and the underlying structural composition.

\noindent\textbf{Line-Guided Region Redrawing} —  
Given a source sketch image $I_s$, a binary mask $M \in \{0,1\}^{H \times W}$ indicating the target region, and user-provided guidance lines $\hat{G}$ within $M$, the objective is to redraw the masked region so that it harmonizes seamlessly with the unmasked context of $I_s$. Concretely, the model must respect the strokes in $\hat{G}$ as explicit structural constraints (e.g., contours, proportions) while matching the global artistic style.

Because these two tasks differ fundamentally in their control modalities (global text vs. local lines), we design a systematic, multi-stage pipeline to construct dedicated, high-quality datasets for each.

\subsection{Instruction-Guided Dataset Construction}

\begin{figure*}[!ht]
  \centering
   \includegraphics[width=1\linewidth]{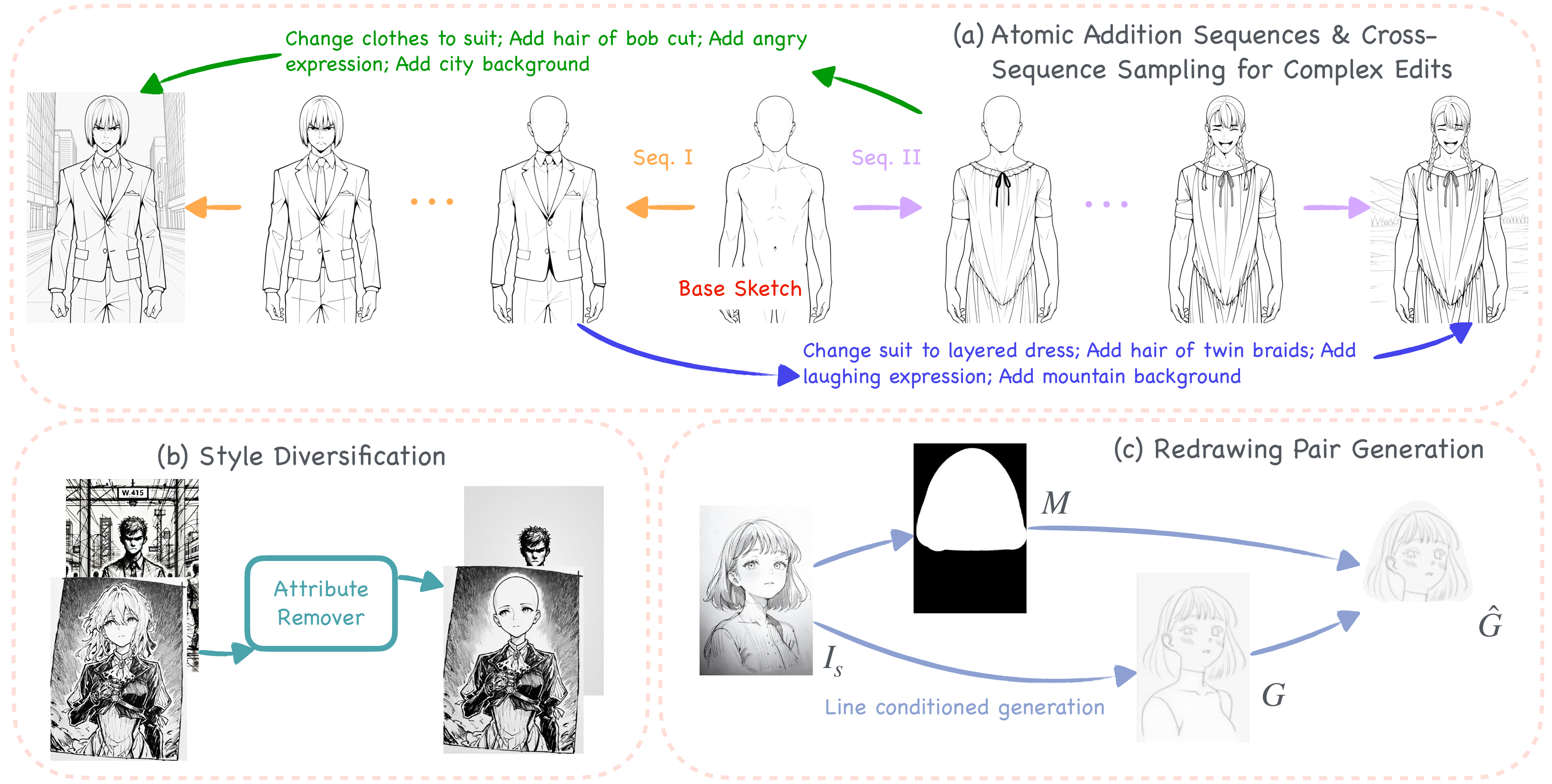}
   \caption{Overview of our data generation pipeline. 
   (a) \textbf{Atomic Addition Sequences \& Cross-Sequence Sampling for Complex Edits} — 
Starting from an attribute-free base sketch, we first generate multiple \textbf{atomic addition sequences} (e.g., Seq. I and Seq. II) by progressively adding single attributes via fixed-seed evolving prompts. 
To break the limits of linear generation, \textbf{Cross-Sequence Sampling} is introduced to pair sketches across different Sequences and time steps. This strategy combinatorially synthesizes \textbf{complex edit pairs} with diverse instructions—including \emph{add}, \emph{remove}, and \emph{replace}—while ensuring perfect structural alignment through their shared base sketch.
   (b) \textbf{Style Diversification} — a specialized removal model, trained on the reverse of our atomic edits, is applied to multi-style sketches to erase specific attributes via Style-Preserving Removal. (c) \textbf{Redrawing Pair Generation} — To simulate user guidance, coarse strokes are generated as structural hints and integrated with the original sketch via target masks. This process constructs training pairs where the model learns to synthesize coherent linework conditioned on both the provided strokes and the unmasked context.}
   \label{fig:data}
\end{figure*}

To train a robust instruction-guided editing model, our dataset construction elegantly addresses three key challenges through a progressive pipeline. First, we establish reliable instruction-image alignment by generating atomic addition sequences (Sec.~\ref{sec:base_sequence}), where each generation step strictly introduces a single attribute. 

Second, to overcome the scarcity of complex editing data, we propose cross-sequence sampling as our core data-synthesis engine (Sec.~\ref{sec:editing_chain}). Rather than being limited to simple, linear modifications, this sampling strategy synthesizes diverse, multi-step edit scenarios while inherently guaranteeing structural alignment, all systematically governed by an edit distance control mechanism. 

Finally, we overcome the stylistic bias of a single generator by expanding the dataset's stylistic coverage via a style-preserving removal strategy (Sec.~\ref{sec:style_div}), which is more robust than addition for cross-style generalization. This unified approach yields a highly diverse, stylistically rich, and precisely controlled training corpus.

\subsubsection{Step 1: Constructing Atomic Addition Sequences}
\label{sec:base_sequence}

In Step~1, we construct coarse-to-fine generation sequences by introducing visual attributes \emph{one at a time}, establishing a foundation of purely additive \textbf{atomic edits}.

As shown in Fig.~\ref{fig:data}(a), we begin with an \emph{attribute-free} base sketch $I^{(0)}$ containing only structural contours (e.g., pose and proportions). For data generation, we utilize Danbooru-style tags and randomly sample an ordered list of attribute tags $\{a_1, a_2, \ldots, a_T\}$ from a vocabulary defined by a categorized WD14 tag~\cite{wd14tagger} set. Let $A^{(t)}$ denote the cumulative attribute set at step $t$, where $A^{(t)} = A^{(t-1)} \cup \{a_t\}$ and $A^{(0)} = \emptyset$.

Starting from $I^{(0)}$, we iteratively synthesize sketches $\{I^{(t)}\}_{t=1}^{T}$ by enforcing evolving prompts (e.g., transitioning from ``bald'' to ``long hair''). To reliably synthesize these attributes while ensuring strict spatial consistency, we utilize fixed generation seeds alongside ControlNet~\cite{Zhang_2023_ICCV} conditioned on the previous sketch $I^{(t-1)}$. 

This progressive generation yields a sequence where each transition $(I^{(t-1)}, I^{(t)})$ constitutes a single-attribute \emph{add} operation---a fundamental \textbf{atomic edit} with an \textbf{edit distance} of $D=1$. By independently repeating this process $M$ times for the same $I^{(0)}$ with different tag sequences, we construct $M$ distinct generation branches. These structurally aligned sequences provide the necessary pool for complex cross-sequence sampling in Step~2.

\subsubsection{Step 2: Synthesizing Complex Edits with Edit Distance Control}
\label{sec:editing_chain}

Building upon the multiple sequences of atomic edits generated in Step~1 (which all branch from the same base sketch), we perform cross-sequence sampling to synthesize diverse and complex edit instructions. We systematically control this synthesis process by quantifying the complexity of each sampled pair using a formal edit distance metric.

Let $I^{(m,t)}$ denote the sketch at step $t$ in the $m$-th sequence, and $A^{(m,t)}$ be its corresponding set of attribute tags. Because all sequences originate from the exact same base sketch $I^{(0)}$, any sampled source sketch $(I_s, A_s)$ and target sketch $(I_t, A_t)$ are inherently aligned in their underlying pose and composition.

We define the semantic edit distance between any source-target pair based on the difference between their attribute sets. This distance is explicitly decomposed into three atomic operations:
\begin{itemize}
    \item \textbf{Add} ($O_{add}$): Attributes present in the target but not in the source, computed as $A_t \setminus A_s$.
    \item \textbf{Remove} ($O_{rm}$): Attributes present in the source but not in the target, computed as $A_s \setminus A_t$.
    \item \textbf{Replace} ($O_{rep}$): Mutually exclusive attributes within the same category (e.g., swapping ``short hair'' for ``long hair'').
\end{itemize}

The total edit distance, denoted as $D = |O_{add}| + |O_{rm}| + |O_{rep}|$, precisely measures the transformation's complexity. By aggregating these atomic operations, we dynamically synthesize comprehensive, multi-attribute instructions. For instance, an instruction like ``remove hat, change to long hair, add scarf'' corresponds to an edit distance of $D=3$. 

This scalable control mechanism allows us to systematically train the model on instructions ranging from simple atomic edits ($D=1$) to highly complex combinations ($D \ge 3$). By strictly controlling the edit distance while preserving the structural layout, our dataset prevents the model from relying on spurious correlations and forces it to learn precise, instruction-following capabilities.

\subsubsection{Step 3: Style Diversification via Style-Preserving Removal}
\label{sec:style_div}

Although datasets from Steps~\ref{sec:base_sequence} and~\ref{sec:editing_chain} cover diverse edit distances, they are limited to a narrow style from a single generator. To expand stylistic coverage, we collect real-world sketches from diverse sources.

Applying our additive pipeline to these sketches often disrupts their coherence, and the ControlNet generator struggles with deletion due to its rigid spatial conditioning. To address this, we train a \textbf{style-preserving removal} model using the reverse of the atomic addition pairs from Step~\ref{sec:base_sequence} (i.e., treating $(I^{(t)}, I^{(t-1)})$ as an atomic removal $O_{rm}$). 

Applying this model to the collected multi-style sketches yields high-quality removal pairs (Fig.~\ref{fig:data}(b)), successfully injecting diverse stylistic priors into our dataset while maintaining precise attribute control.

\subsection{Line-Guided Redrawing Dataset Construction}
\label{sec:redrawing_data}

The dataset for line-guided region redrawing combines two sources: sketches from Step~\ref{sec:base_sequence} and the diverse sketches from Step~\ref{sec:style_div}. This ensures comprehensive coverage of both synthetic and real-world artistic styles.

To simulate diverse user guidance, we rely on an Anime Lineart preprocessor to extract linework from each sketch $I_s$, and utilize a ControlNet-based diffusion model to generate varied guidance line images $G$ (Fig.~\ref{fig:data}(c)). These generated lines vary in detail and coverage, and may exhibit imperfections, effectively mimicking realistic hand-drawn strokes.

To cover diverse redrawing scenarios, we construct binary masks $M$ in two forms: (1) semantically meaningful masks derived from body-part and object detection, and (2) randomly generated masks of arbitrary shapes.

Formally, we construct input-target training tuples $(I_s, M, G, \hat{G})$, where $I_s$ is the original target sketch, $M$ is the binary mask, $G$ is the full synthesized guidance image, and $\hat{G}$ denotes the masked guidance region (i.e., $\hat{G} = G \odot M$). During training, the model is tasked with redrawing $\hat{G}$ to perfectly harmonize with the unmasked context of $I_s$. This robust formulation covers a broad spectrum of editing cases, ranging from loosely defined regions to semantically precise local modifications.

\section{Method}
\label{sec:method}

\begin{figure*}[!ht]
  \centering
   \includegraphics[width=0.95\linewidth]{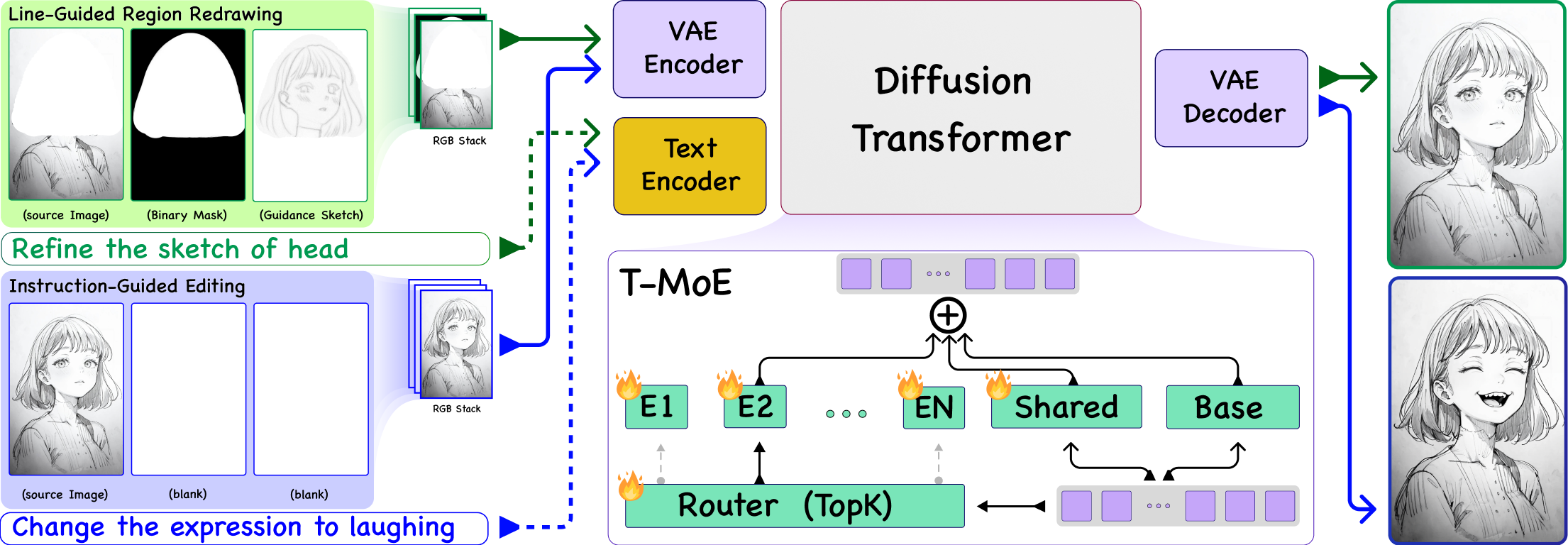}
   \caption{Overview of our unified SketchAssist framework. It seamlessly handles both (a) Line-Guided Region Redrawing (conditioned on a source, mask, and guidance lines) and (b) Instruction-Guided Editing (conditioned on text and a blank global mask). To process these distinct modalities efficiently, our Diffusion Transformer incorporates a Task-guided Mixture-of-Experts (T-MoE) module within its LoRA layers, dynamically routing visual-text features to task-specific experts.}
   \label{fig:model}
\end{figure*}

To support both instruction-guided editing and line-guided redrawing within a single diffusion transformer (DiT), we propose a unified framework (Fig.~\ref{fig:model}) with minimal architectural overhead. It relies on two core designs: a unified input conditioning strategy to flexibly switch between editing modes, and a Task-guided Mixture-of-Experts (T-MoE) module to adapt network parameters to distinct task behaviors.

\subsection{Unified Input Condition Construction}
\label{sec:tmoe}

To incorporate localized redrawing into an instruction-based editing model, existing approaches typically expand the input channel dimension or add separate control branches (e.g., OminiControl~\cite{tan2025ominicontrol}), which increases training and inference overhead. Leveraging the monochrome nature of sketches, we circumvent this by elegantly packing all spatial conditions into a standard 3-channel composite image $\mathcal{I}_{cond} \in \mathbb{R}^{H \times W \times 3}$.

We use a unified multi-channel representation implemented as three RGB channels to encode source sketches, editable regions, and structural guidance.

\smallskip\noindent
\textbf{R Channel (Source Context):} Encodes the base sketch to be edited. \textit{Instruction:} The complete original sketch. \textit{Line:} The original sketch with the masked target region erased.

\smallskip\noindent
\textbf{G Channel (Editable Mask):} Encodes the binary mask $M$, where white (1) denotes editable regions (e.g., the `blank' global mask in Fig.~\ref{fig:model}) and black (0) denotes preserved context. \textit{Instruction:} A global mask for holistic attribute changes. \textit{Line:} A localized mask for regional modifications.

\smallskip\noindent
\textbf{B Channel (Structural Guidance):} Encodes the line-based visual guidance $G$. \textit{Instruction:} An empty canvas, as edits are solely guided by the text instruction $t$. \textit{Line:} The user-provided strokes or extracted guidance lines within the masked region.

By explicitly routing task conditions through this structured 3-channel format, the model seamlessly switches between global semantic edits and precise local redrawing without requiring extra pipelines or input layer retraining.

\subsection{Task-guided Mixture-of-Experts (T-MoE)}

While our unified input handles spatial conditions, the two tasks require distinct feature mappings. To decouple these behaviors without inflating the model size, we introduce a Task-guided Mixture-of-Experts (T-MoE) architecture integrated into the LoRA~\cite{hu2022lora} layers. 

Each T-MoE layer consists of a \emph{Shared LoRA} module to capture task-agnostic structural and stylistic priors, and a set of \emph{Expert LoRA} modules to learn task-specific editing patterns. During inference, a routing input $z$---formed by concatenating text and visual features---dynamically selects the most relevant experts. The final output is computed as:

\begin{equation}
\begin{aligned}
    \text{Output} &= \text{BaseLayer}(x) + \mathrm{SharedLoRA}(x) \\
    &\quad + \frac{\alpha}{r} \sum_{i=1}^{N} G(z)_i \cdot \mathrm{ExpertLoRA}_i(x),
\end{aligned}
\end{equation}
where $\alpha$ and $r$ are the LoRA scaling factor and rank. The routing probability for the $i$-th expert is $G(z)_i = \mathrm{Softmax}\big( \mathrm{TopK}\big( g(z), K \big) \big)_i$, which ensures sparse activation by retaining only the top-$K$ entries from the pre-softmax routing scores $g(z)$. This design enables highly specialized task execution while maintaining a shared foundation for sketch coherence.

\section{Experiments}
\label{sec:experiments}

\subsection{Implementation Details}

\textbf{Dataset Construction \& Filtering.} For instruction-guided editing, we generate 10,000 distinct base sketches, each extended into multiple addition branches. To guarantee high data quality, we employ a rigorous dual-stage filtering pipeline: (1) \textit{Stability}: we use Human-Art~\cite{ju2023human} and CLIP~\cite{Clip} to enforce strict pose and style consistency; (2) \textit{Semantics}: we utilize WD14 Tagger~\cite{wd14tagger} to confirm the precise presence/absence of target attributes, and Qwen-VL to validate overall instruction alignment. This strict filtering yields approximately 100,000 high-quality training images. For style diversification, we collect 4,000 multi-style artworks, identify their existing attributes via WD14 Tagger, and apply our removal model guided by constructed instructions (e.g., ``remove [tag]''). For line-guided redrawing, binary masks are generated using Grounding DINO~\cite{liu2024grounding} and SAM2~\cite{ravi2024sam} for object-level regions, while body-level masks are derived from stochastic regions around Human-Art keypoints to simulate imprecise, realistic user strokes. 

\textbf{Training Details.} We train our unified framework based on the FLUX.1-Kontext~\cite{FLUX} architecture. While data generation relies on predefined tags, the final model is trained entirely on natural language instructions to ensure user-friendly interaction. During training, we employ the cross-sequence sampling strategy to dynamically form edit pairs with an edit distance ranging from $D=1$ to $5$, simulating both atomic and complex multi-step workflows. Comprehensive details regarding model configuration, T-MoE active-parameter alignment (designed for fair comparison against standard LoRA baselines), and  hyper-parameters are provided in the supplementary material.

\subsection{Quantitative Evaluation}

\begin{table*}[t]
\centering
\setlength{\tabcolsep}{4pt}
\begin{tabular}{lcccccccccc}
\hline
Model & CLIP-T$\uparrow$ & CLIP-I$\uparrow$ & DINO$\uparrow$ & Q\_SC$\uparrow$ & Q\_PQ$\uparrow$ & Q\_O$\uparrow$ & G\_SC$\uparrow$ & G\_PQ$\uparrow$ & G\_O$\uparrow$ & WR(\%) \\
\hline
ICEdit~\cite{Enabling}        & 0.270 & 0.811 & 0.705 & 4.13 & 7.14   & 4.54 & 5.86 & 7.13 & 5.98 & 94.81\\ 
FLUX.1 Kontext~\cite{FLUX}      & 0.296 & 0.882 & 0.791 & 6.62 & 7.53   & 6.42 & 7.54 & 7.94 & 7.40 & 84.45 \\ 
Step1X-Edit~\cite{liu2025step1x}      & 0.295 & 0.886 & 0.809 & 7.60 & 7.24   & 7.25 & 8.44 & 7.55 & 7.89 & 82.76 \\ 
Qwen-Image-Edit~\cite{Qwen}     & 0.292 & 0.849 & 0.776 & 6.89 & \textbf{7.66}   & 6.79 & 8.13 & 7.87 & 7.89 & 86.91 \\ 
\textbf{Ours}                         & \textbf{0.305} & \textbf{0.931} & \textbf{0.879} & \textbf{8.74} & 7.49 & \textbf{8.04} & \textbf{8.97} & \textbf{8.17} & \textbf{8.52} & - \\ 
\hline
\end{tabular}
\caption{Quantitative results for Instruction-guided Sketch Editing. }

\label{tab:instr_edit}
\end{table*}
\begin{table}[t]
\centering
\setlength{\tabcolsep}{4pt}
\begin{tabular}{lcccc}
\hline
Model & LPIPS$\downarrow$ & CLIP-I$\uparrow$ & DINO$\uparrow$ & WR(\%) \\
\hline
SketchEdit~\cite{Sketchedit}  & 0.2277 & 0.853 & 0.800 &  98.56 \\ 
BrushNet~\cite{Brushnet}        & 0.1877  & 0.852 & 0.795 & 97.23 \\ 
MagicQuill~\cite{Magicquill}       & 0.1472  & 0.903 & 0.887 & 94.49 \\ 
\textbf{Ours}                         & \textbf{0.0972}  & \textbf{0.961} & \textbf{0.949} & - \\ 
\hline
\end{tabular}
\caption{Quantitative results for Line-guided Region Redrawing. }
\label{tab:region_repaint}
\end{table}

We evaluate our unified framework on separate test sets for the two tasks, each containing 200 curated sketches. To ensure a rigorous evaluation, the instruction-guided samples are generated via the pipeline in Section~\ref{sec:data} using 200 unique base sketches not seen during training. We utilize diverse prompts from Gemini covering various styles, attributes, and backgrounds, followed by manual review for accuracy and clarity. The line-guided samples are produced with manual masks defining target regions, complemented by quality control to ensure semantic validity. Since all test data are pipeline-generated, we have access to accurate target images and ground-truth captions for precise metric calculation.

For the instruction-guided task, we compute CLIP-T~\cite{Clip}, CLIP-I, and DINOv2~\cite{oquab2023dinov2} scores, alongside SC (Semantic Consistency), PQ (Perceptual Quality), and O (Overall Score) following the VIEScore~\cite{ku2024viescore} protocol powered by Gemini-2.0-Flash and Qwen3-VL-30B~\cite{bai2025qwen3}.

For the line-guided task, we report LPIPS~\cite{lpips}, CLIP-I, and DINO scores. Pixel-level metrics like PSNR and SSIM~\cite{ssim} are provided in the supplementary material for reference, as they can be biased by the large white background of sketches.

Table~\ref{tab:instr_edit} and Table~\ref{tab:region_repaint} report the quantitative results. Our method outperforms baselines across the majority of metrics, reflecting superior instruction adherence and stylistic fidelity. While Qwen-Image-Edit~\cite{Qwen} attains a high Q\_PQ score, our method achieves the highest Overall score, demonstrating a better balance between semantic accuracy and visual coherence.

\noindent\textbf{User Study.} 
To further assess human preference, we conducted a formal user study on an in-house crowdsourcing platform involving 50 raters. In each trial, raters performed side-by-side comparisons between our results and baselines, selecting the superior output based on instruction adherence, structural integrity, and style consistency. Our approach achieves a markedly high winning rate (WR) against all competing baselines, as shown in the tables.

\begin{table*}[t]
\centering
\begin{tabular}{l|cccccc|cccc}
\hline
 & \multicolumn{6}{c|}{Instruction-guided Editing} 
 & \multicolumn{3}{c}{Line-Guided Region Redrawing} \\
\cline{2-10}
Model Variant 
& CLIP-T  & CLIP-I & DINO & Q\_SC & Q\_PQ & Q\_O 
& LPIPS &  CLIP-I & DINO \\
\hline
Baseline & 0.296 & 0.882 & 0.791 & 6.62 & 7.53 & 6.42 &  0.119 & 0.924 & 0.890 \\ 
+ Atomic Addition Seq. & 0.288 & 0.892 & 0.831 & 7.31 & 7.67 & 7.20 &  0.119 & 0.946 & 0.937 \\ 
+ Cross-Sequence Sampling       & 0.296 & 0.901 & 0.847 & 7.92 & \textbf{7.67} & 7.55 &  0.112 & 0.950 & 0.948 \\ 
+ Style Diversity  & 0.301 & 0.931 & 0.879 & 8.56 & 7.44 & 7.86 & 0.0915 & 0.956 & 0.956 \\ 
+ T-MoE (full)       & \textbf{0.305} & \textbf{0.931} & \textbf{0.879} & \textbf{8.74} & 7.49 & \textbf{8.04} &  \textbf{0.0904} & \textbf{0.960} & \textbf{0.956} \\ 
\hline
\end{tabular}
\caption{
Ablation study on Instruction-guided  Editing (left) and Line-Guided Region Redrawing (right) under progressive module addition. 
}
\label{tab:ablation_combined}
\end{table*}

\subsection{Qualitative Comparison}

We present qualitative comparisons for both tasks against state-of-the-art approaches.

\begin{figure*}[!ht]
  \centering
   \includegraphics[width=0.95\linewidth]{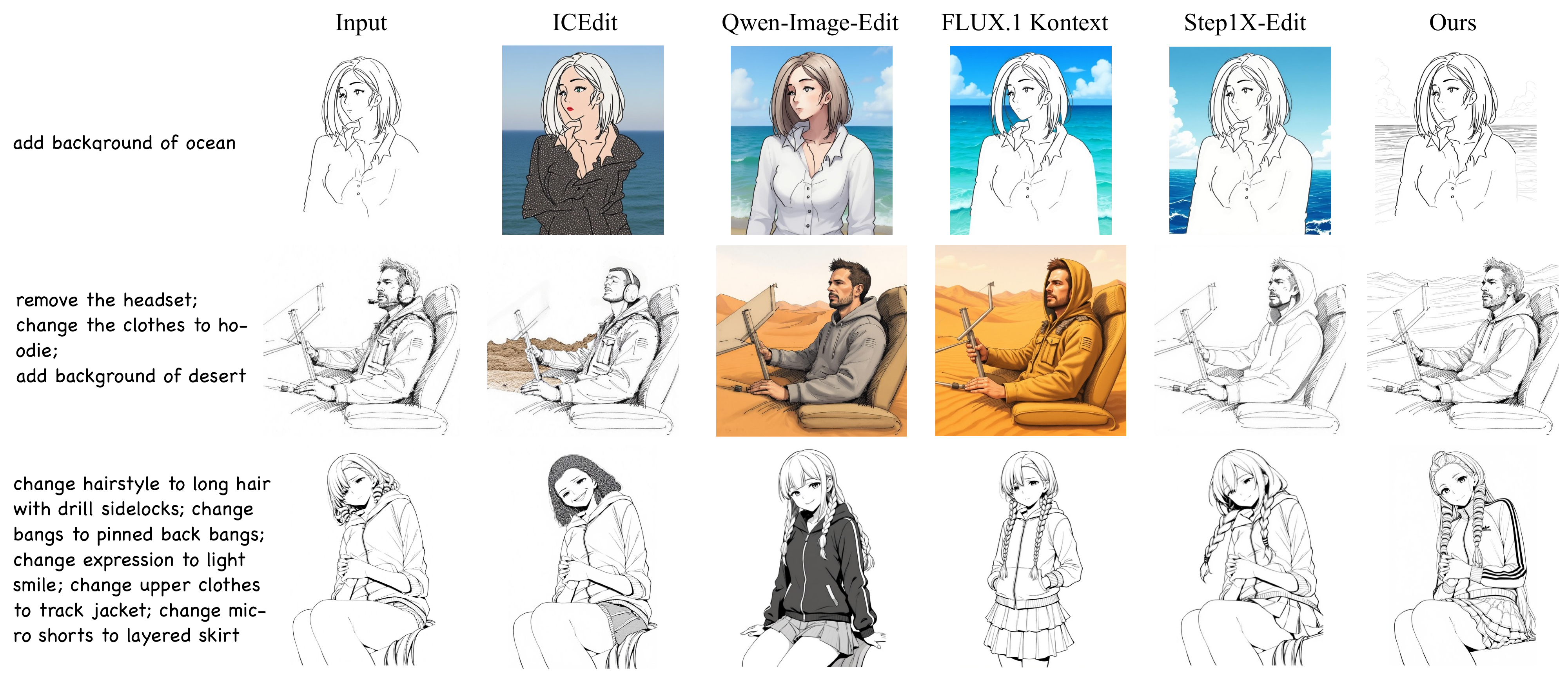}
   \caption{Qualitative comparisons of Instruction-guided Sketch Editing}
   \label{fig:qua_comp1}
\end{figure*}

For the Instruction-guided Sketch Editing task (Fig.~\ref{fig:qua_comp1}), competing models often introduce substantial modifications that deviate from the instructions or fail to maintain the original style. In contrast, our method accurately applies requested edits while preserving structural and stylistic fidelity.Notably, even under complex, multi-step instructions (e.g., the bottom case), SketchAssist ensures the stability of the sketch's identity, whereas other models often suffer from content drift or inaccurate edits as the number of operations increases.

\begin{figure*}[!ht]
  \centering
   \includegraphics[width=0.95\linewidth]{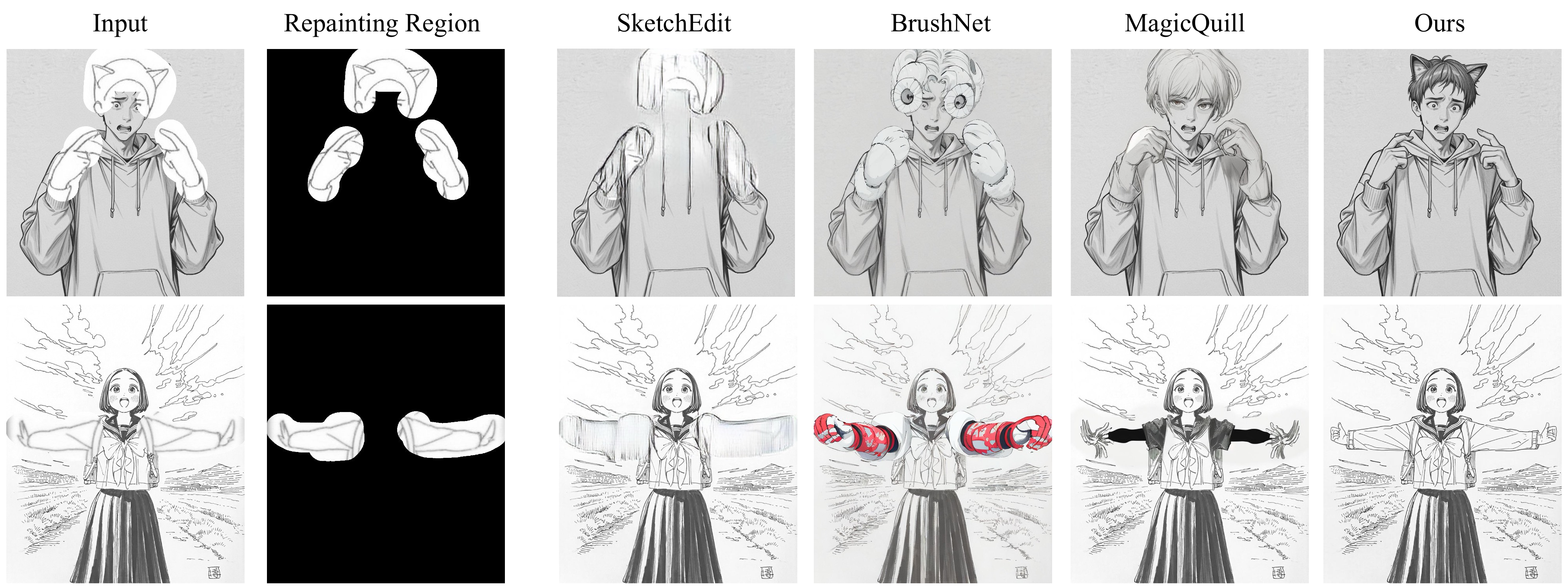}
   \caption{Qualitative comparisons of Line-guided Region Redrawing}
   \label{fig:qua_comp2}
\end{figure*}

For the Line-guided Region Redrawing task (Fig.~\ref{fig:qua_comp2}), SketchEdit and BrushNet fail to consistently utilize the provided line guidance, producing modifications that diverge from the specified regions or ignore fine contour details. MagicQuill generates outputs that are stylistically close to the target, yet the redrawn regions frequently fail to align with the  guidance lines. Ours adheres closely to the provided guidance lines and produces clean, well-integrated modifications that align with the original sketch structure.

These visual observations are consistent with the quantitative results, further demonstrating that SketchAssist achieves superior instruction adherence, structural accuracy, and style preservation across both tasks.


\subsection{Ablation Study}

Table~\ref{tab:ablation_combined} illustrates the progressive performance gains achieved by integrating our proposed modules. To rigorously evaluate the structural robustness of SketchAssist, we expand the redrawing test set to 400 samples. This includes the original 200 semantic-mask pairs and 200 additional samples using random masks without semantic definitions. This augmented set stress-tests the model's versatility in scenarios ranging from instruction-driven local edits to pure line-guided structural refinement where textual guidance is absent.

\noindent\textbf{Impact of Chained Data Synthesis.} 
The Atomic Addition Sequence (Sec.~\ref{sec:base_sequence}) establishes a stable baseline for single-attribute edits. Building upon this, Cross-Sequence Sampling (Sec.~\ref{sec:editing_chain}) introduces complex, multi-step edit scenarios ($D=1 \sim 5$). This effectively improves structural consistency (CLIP-I: +0.009; DINO: +0.016), as it forces the model to maintain the subject's identity across diverse attribute combinations.

\noindent\textbf{Impact of Style Diversification.} 
Integrating multi-style data via Style-Preserving Removal (Sec.~\ref{sec:style_div}) significantly enhances semantic and stylistic alignment across both tasks. We observe a marginal decrease in PQ scores; this is a fidelity-quality trade-off, as the model faithfully preserves the raw, hand-drawn textures of diverse source sketches rather than over-smoothing them into generic, ``clean'' line art.

\noindent\textbf{Effect of T-MoE.} 
The T-MoE (Sec.~\ref{sec:tmoe}) module achieves the best overall performance across nearly all metrics. By dynamically routing task-specific features, it effectively mitigates parameter interference between global semantic editing and local structural redrawing. This specialization allows a single DiT to maintain high style fidelity while executing precise, localized modifications.

\section{Conclusion}
\label{conclusion}
We present \textbf{SketchAssist}, an interactive sketch drawing assistant that unifies instruction-guided editing and line-guided region redrawing for practical, coarse-to-fine creation. On the data side, we introduce a controllable pipeline that constructs attribute-edit sequences from attribute-free bases, forms multi-step edit chains, and expands stylistic diversity. On the model side, we use RGB channels as a unified input representation to encode sketches, masks, and guidance signals, and integrate a task-guided Mixture-of-Experts into LoRA to adapt across editing modes while preserving structure and style. Experiments demonstrate state-of-the-art performance on both instruction-guided and line-guided tasks, with strong instruction adherence, structural fidelity, and stylistic consistency.

{
    \small
    \bibliographystyle{ieeenat_fullname}
    \bibliography{main}

@String(CVPR= {IEEE Conf. Comput. Vis. Pattern Recog.})

@String(ICCV= {Int. Conf. Comput. Vis.})

@String(ICLR = {Int. Conf. Learn. Represent.})

@String(CVPR  = {CVPR})

@String(ICCV  = {ICCV})

@String(ICLR  = {ICLR})

@InProceedings{Zhang_2023_ICCV,
    author    = {Zhang, Lvmin and Rao, Anyi and Agrawala, Maneesh},
    title     = {Adding Conditional Control to Text-to-Image Diffusion Models},
    booktitle = {Proceedings of the IEEE/CVF International Conference on Computer Vision (ICCV)},
    month     = {October},
    year      = {2023},
    pages     = {3836-3847}
}

@article{liu2025step1x,
  title={Step1x-edit: A practical framework for general image editing},
  author={Liu, Shiyu and Han, Yucheng and Xing, Peng and Yin, Fukun and Wang, Rui and Cheng, Wei and Liao, Jiaqi and Wang, Yingming and Fu, Honghao and Han, Chunrui and others},
  journal={arXiv preprint arXiv:2504.17761},
  year={2025}
}

@inproceedings{ku2024viescore,
  title={Viescore: Towards explainable metrics for conditional image synthesis evaluation},
  author={Ku, Max and Jiang, Dongfu and Wei, Cong and Yue, Xiang and Chen, Wenhu},
  booktitle={Proceedings of the 62nd Annual Meeting of the Association for Computational Linguistics (Volume 1: Long Papers)},
  pages={12268--12290},
  year={2024}
}

@article{Generative,
  title={Generative adversarial nets},
  author={Goodfellow, Ian J and Pouget-Abadie, Jean and Mirza, Mehdi and Xu, Bing and Warde-Farley, David and Ozair, Sherjil and Courville, Aaron and Bengio, Yoshua},
  journal={Advances in neural information processing systems},
  volume={27},
  year={2014}
}

@article{Large,
  title={Large scale GAN training for high fidelity natural image synthesis},
  author={Brock, Andrew and Donahue, Jeff and Simonyan, Karen},
  journal={arXiv preprint arXiv:1809.11096},
  year={2018}
}

@article{Alias-free,
  title={Alias-free generative adversarial networks},
  author={Karras, Tero and Aittala, Miika and Laine, Samuli and H{\"a}rk{\"o}nen, Erik and Hellsten, Janne and Lehtinen, Jaakko and Aila, Timo},
  journal={Advances in neural information processing systems},
  volume={34},
  pages={852--863},
  year={2021}
}

@inproceedings{A-style-based-generator,
  title={A style-based generator architecture for generative adversarial networks},
  author={Karras, Tero and Laine, Samuli and Aila, Timo},
  booktitle={Proceedings of the IEEE/CVF conference on computer vision and pattern recognition},
  pages={4401--4410},
  year={2019}
}

@inproceedings{Analyzing,
  title={Analyzing and improving the image quality of stylegan},
  author={Karras, Tero and Laine, Samuli and Aittala, Miika and Hellsten, Janne and Lehtinen, Jaakko and Aila, Timo},
  booktitle={Proceedings of the IEEE/CVF conference on computer vision and pattern recognition},
  pages={8110--8119},
  year={2020}
}

@inproceedings{Clip,
  title={Learning transferable visual models from natural language supervision},
  author={Radford, Alec and Kim, Jong Wook and Hallacy, Chris and Ramesh, Aditya and Goh, Gabriel and Agarwal, Sandhini and Sastry, Girish and Askell, Amanda and Mishkin, Pamela and Clark, Jack and others},
  booktitle={International conference on machine learning},
  pages={8748--8763},
  year={2021},
  organization={PmLR}
}

@inproceedings{Vqgan-clip,
  title={Vqgan-clip: Open domain image generation and editing with natural language guidance},
  author={Crowson, Katherine and Biderman, Stella and Kornis, Daniel and Stander, Dashiell and Hallahan, Eric and Castricato, Louis and Raff, Edward},
  booktitle={European conference on computer vision},
  pages={88--105},
  year={2022},
  organization={Springer}
}

@inproceedings{Taming,
  title={Taming transformers for high-resolution image synthesis},
  author={Esser, Patrick and Rombach, Robin and Ommer, Bjorn},
  booktitle={Proceedings of the IEEE/CVF conference on computer vision and pattern recognition},
  pages={12873--12883},
  year={2021}
}

@inproceedings{Blended,
  title={Blended diffusion for text-driven editing of natural images},
  author={Avrahami, Omri and Lischinski, Dani and Fried, Ohad},
  booktitle={Proceedings of the IEEE/CVF conference on computer vision and pattern recognition},
  pages={18208--18218},
  year={2022}
}

@article{Denoising,
  title={Denoising diffusion probabilistic models},
  author={Ho, Jonathan and Jain, Ajay and Abbeel, Pieter},
  journal={Advances in neural information processing systems},
  volume={33},
  pages={6840--6851},
  year={2020}
}

@inproceedings{SD,
  title={High-resolution image synthesis with latent diffusion models},
  author={Rombach, Robin and Blattmann, Andreas and Lorenz, Dominik and Esser, Patrick and Ommer, Bj{\"o}rn},
  booktitle={Proceedings of the IEEE/CVF conference on computer vision and pattern recognition},
  pages={10684--10695},
  year={2022}
}

@article{FLUX,
  title={FLUX. 1 Kontext: Flow Matching for In-Context Image Generation and Editing in Latent Space},
  author={Labs, Black Forest and Batifol, Stephen and Blattmann, Andreas and Boesel, Frederic and Consul, Saksham and Diagne, Cyril and Dockhorn, Tim and English, Jack and English, Zion and Esser, Patrick and others},
  journal={arXiv preprint arXiv:2506.15742},
  year={2025}
}

@article{Qwen,
  title={Qwen-image technical report},
  author={Wu, Chenfei and Li, Jiahao and Zhou, Jingren and Lin, Junyang and Gao, Kaiyuan and Yan, Kun and Yin, Sheng-ming and Bai, Shuai and Xu, Xiao and Chen, Yilei and others},
  journal={arXiv preprint arXiv:2508.02324},
  year={2025}
}

@article{bai2025qwen3,
  title={Qwen3-vl technical report},
  author={Bai, Shuai and Cai, Yuxuan and Chen, Ruizhe and Chen, Keqin and Chen, Xionghui and Cheng, Zesen and Deng, Lianghao and Ding, Wei and Gao, Chang and Ge, Chunjiang and others},
  journal={arXiv preprint arXiv:2511.21631},
  year={2025}
}

@inproceedings{Enabling,
  title={Enabling Instructional Image Editing with In-Context Generation in Large Scale Diffusion Transformer},
  author={Zhang, Zechuan and Xie, Ji and Lu, Yu and Yang, Zongxin and Yang, Yi},
  booktitle={The Thirty-ninth Annual Conference on Neural Information Processing Systems}
}

@inproceedings{Dit,
  title={Scalable diffusion models with transformers},
  author={Peebles, William and Xie, Saining},
  booktitle={Proceedings of the IEEE/CVF international conference on computer vision},
  pages={4195--4205},
  year={2023}
}

@inproceedings{ju2023human,
  title={Human-art: A versatile human-centric dataset bridging natural and artificial scenes},
  author={Ju, Xuan and Zeng, Ailing and Wang, Jianan and Xu, Qiang and Zhang, Lei},
  booktitle={Proceedings of the IEEE/CVF conference on computer vision and pattern recognition},
  pages={618--629},
  year={2023}
}

@inproceedings{Repaint,
  title={Repaint: Inpainting using denoising diffusion probabilistic models},
  author={Lugmayr, Andreas and Danelljan, Martin and Romero, Andres and Yu, Fisher and Timofte, Radu and Van Gool, Luc},
  booktitle={Proceedings of the IEEE/CVF conference on computer vision and pattern recognition},
  pages={11461--11471},
  year={2022}
}

@inproceedings{Brushnet,
  title={Brushnet: A plug-and-play image inpainting model with decomposed dual-branch diffusion},
  author={Ju, Xuan and Liu, Xian and Wang, Xintao and Bian, Yuxuan and Shan, Ying and Xu, Qiang},
  booktitle={European Conference on Computer Vision},
  pages={150--168},
  year={2024},
  organization={Springer}
}

@inproceedings{Text2live,
  title={Text2live: Text-driven layered image and video editing},
  author={Bar-Tal, Omer and Ofri-Amar, Dolev and Fridman, Rafail and Kasten, Yoni and Dekel, Tali},
  booktitle={European conference on computer vision},
  pages={707--723},
  year={2022},
  organization={Springer}
}

@article{Region-Prompt,
  title={Training-free regional prompting for diffusion transformers},
  author={Chen, Anthony and Xu, Jianjin and Zheng, Wenzhao and Dai, Gaole and Wang, Yida and Zhang, Renrui and Wang, Haofan and Zhang, Shanghang},
  journal={arXiv preprint arXiv:2411.02395},
  year={2024}
}

@inproceedings{Sketchedit,
  title={Sketchedit: Mask-free local image manipulation with partial sketches},
  author={Zeng, Yu and Lin, Zhe and Patel, Vishal M},
  booktitle={Proceedings of the IEEE/CVF conference on computer vision and pattern recognition},
  pages={5951--5961},
  year={2022}
}

@inproceedings{Magicquill,
  title={Magicquill: An intelligent interactive image editing system},
  author={Liu, Zichen and Yu, Yue and Ouyang, Hao and Wang, Qiuyu and Cheng, Ka Leong and Wang, Wen and Liu, Zhiheng and Chen, Qifeng and Shen, Yujun},
  booktitle={Proceedings of the Computer Vision and Pattern Recognition Conference},
  pages={13072--13082},
  year={2025}
}

@inproceedings{tan2025ominicontrol,
  title={Ominicontrol: Minimal and universal control for diffusion transformer},
  author={Tan, Zhenxiong and Liu, Songhua and Yang, Xingyi and Xue, Qiaochu and Wang, Xinchao},
  booktitle={Proceedings of the IEEE/CVF International Conference on Computer Vision},
  pages={14940--14950},
  year={2025}
}

@inproceedings{lpips,
  title={The unreasonable effectiveness of deep features as a perceptual metric},
  author={Zhang, Richard and Isola, Phillip and Efros, Alexei A and Shechtman, Eli and Wang, Oliver},
  booktitle={Proceedings of the IEEE conference on computer vision and pattern recognition},
  pages={586--595},
  year={2018}
}

@article{oquab2023dinov2,
  title={Dinov2: Learning robust visual features without supervision},
  author={Oquab, Maxime and Darcet, Timoth{\'e}e and Moutakanni, Th{\'e}o and Vo, Huy and Szafraniec, Marc and Khalidov, Vasil and Fernandez, Pierre and Haziza, Daniel and Massa, Francisco and El-Nouby, Alaaeldin and others},
  journal={arXiv preprint arXiv:2304.07193},
  year={2023}
}

@article{ssim,
  title={Image quality assessment: from error visibility to structural similarity},
  author={Wang, Zhou and Bovik, Alan C and Sheikh, Hamid R and Simoncelli, Eero P},
  journal={IEEE transactions on image processing},
  volume={13},
  number={4},
  pages={600--612},
  year={2004},
  publisher={IEEE}
}

@article{ravi2024sam,
  title={Sam 2: Segment anything in images and videos},
  author={Ravi, Nikhila and Gabeur, Valentin and Hu, Yuan-Ting and Hu, Ronghang and Ryali, Chaitanya and Ma, Tengyu and Khedr, Haitham and R{\"a}dle, Roman and Rolland, Chloe and Gustafson, Laura and others},
  journal={arXiv preprint arXiv:2408.00714},
  year={2024}
}

@article{hu2022lora,
  title={Lora: Low-rank adaptation of large language models.},
  author={Hu, Edward J and Shen, Yelong and Wallis, Phillip and Allen-Zhu, Zeyuan and Li, Yuanzhi and Wang, Shean and Wang, Lu and Chen, Weizhu and others},
  journal={ICLR},
  volume={1},
  number={2},
  pages={3},
  year={2022}
}

@article{hertz2022prompt,
  title={Prompt-to-prompt image editing with cross attention control},
  author={Hertz, Amir and Mokady, Ron and Tenenbaum, Jay and Aberman, Kfir and Pritch, Yael and Cohen-Or, Daniel},
  journal={arXiv preprint arXiv:2208.01626},
  year={2022}
}

@inproceedings{kawar2023imagic,
  title={Imagic: Text-based real image editing with diffusion models},
  author={Kawar, Bahjat and Zada, Shiran and Lang, Oran and Tov, Omer and Chang, Huiwen and Dekel, Tali and Mosseri, Inbar and Irani, Michal},
  booktitle={Proceedings of the IEEE/CVF conference on computer vision and pattern recognition},
  pages={6007--6017},
  year={2023}
}

@inproceedings{brooks2023instructpix2pix,
  title={Instructpix2pix: Learning to follow image editing instructions},
  author={Brooks, Tim and Holynski, Aleksander and Efros, Alexei A},
  booktitle={Proceedings of the IEEE/CVF conference on computer vision and pattern recognition},
  pages={18392--18402},
  year={2023}
}

@InProceedings{Sheynin_2024_CVPR,
    author    = {Sheynin, Shelly and Polyak, Adam and Singer, Uriel and Kirstain, Yuval and Zohar, Amit and Ashual, Oron and Parikh, Devi and Taigman, Yaniv},
    title     = {Emu Edit: Precise Image Editing via Recognition and Generation Tasks},
    booktitle = {Proceedings of the IEEE/CVF Conference on Computer Vision and Pattern Recognition (CVPR)},
    month     = {June},
    year      = {2024},
    pages     = {8871-8879}
}

@inproceedings{yu2025anyedit,
  title={Anyedit: Mastering unified high-quality image editing for any idea},
  author={Yu, Qifan and Chow, Wei and Yue, Zhongqi and Pan, Kaihang and Wu, Yang and Wan, Xiaoyang and Li, Juncheng and Tang, Siliang and Zhang, Hanwang and Zhuang, Yueting},
  booktitle={Proceedings of the Computer Vision and Pattern Recognition Conference},
  pages={26125--26135},
  year={2025}
}

@inproceedings{liu2024grounding,
  title={Grounding dino: Marrying dino with grounded pre-training for open-set object detection},
  author={Liu, Shilong and Zeng, Zhaoyang and Ren, Tianhe and Li, Feng and Zhang, Hao and Yang, Jie and Jiang, Qing and Li, Chunyuan and Yang, Jianwei and Su, Hang and others},
  booktitle={European conference on computer vision},
  pages={38--55},
  year={2024},
  organization={Springer}
}

@InProceedings{Bandyopadhyay_2024_CVPR,
    author    = {Bandyopadhyay, Hmrishav and Bhunia, Ayan Kumar and Chowdhury, Pinaki Nath and Sain, Aneeshan and Xiang, Tao and Hospedales, Timothy and Song, Yi-Zhe},
    title     = {SketchINR: A First Look into Sketches as Implicit Neural Representations},
    booktitle = {Proceedings of the IEEE/CVF Conference on Computer Vision and Pattern Recognition (CVPR)},
    month     = {June},
    year      = {2024},
    pages     = {12565-12574}
}

@article{podell2023sdxl,
  title={Sdxl: Improving latent diffusion models for high-resolution image synthesis},
  author={Podell, Dustin and English, Zion and Lacey, Kyle and Blattmann, Andreas and Dockhorn, Tim and M{\"u}ller, Jonas and Penna, Joe and Rombach, Robin},
  journal={arXiv preprint arXiv:2307.01952},
  year={2023}
}

@article{jintp,
  title={TP-Blend: Textual-Prompt Attention Pairing for Precise Object-Style Blending in Diffusion Models},
  author={Jin, Xin and Zhong, Yichuan and Tian, Yapeng},
  journal={Transactions on Machine Learning Research}
}

@misc{wd14tagger,
  author = {SmilingWolf},
  title = {WD 14 Tagger v3 (EVA02 Large)},
  year = {2024},
  publisher = {Hugging Face},
  howpublished = {\url{https://huggingface.co/SmilingWolf/wd-eva02-large-tagger-v3}},
}
}

\clearpage
\setcounter{page}{1}
\maketitlesupplementary

\section{Detailed Evaluation Protocol}

In this section, we provide further details on the evaluation settings, including the specific prompt configurations used for MLLM-based automated evaluation (VIEScore~\cite{ku2024viescore}), supplementary pixel-wise metrics, and the rationale behind the evaluation scope for different experimental tables.

\subsection{Supplementary Pixel-wise Evaluation}

Complementing the perceptual and semantic evaluations, we additionally provide pixel-wise evaluations (PSNR and SSIM~\cite{ssim}) in this section as supplementary metrics. While pixel-wise metrics are known to correlate poorly with semantic editing quality in sparse sketch domains due to background dominance and sensitivity to spatial misalignments~\cite{lpips,Bandyopadhyay_2024_CVPR}, we include them only to offer a reference for the model’s precision in low-level reconstruction, rather than as primary indicators of overall performance.
\begin{itemize}
    \item \textbf{Global Metrics (PSNR/SSIM):} These provide a coarse proxy for \textbf{structural fidelity in unmasked regions}, loosely reflecting whether the model preserves the original context and blends the edited region without introducing obvious artifacts.
    \item \textbf{Masked-Region Metrics (PSNR-M/SSIM-M):} These offer a limited indication of \textbf{control precision}. Since the guidance lines in our test set are derived from the ground truth, higher scores within the masked region may suggest that the model is able to follow the provided guidance and reconstruct geometry consistent with the target, although such pixel-wise metrics remain imperfect for evaluating semantic editing quality.
\end{itemize}
Our method achieves state-of-the-art results across both dimensions, demonstrating its ability to perform precise local reconstruction while maintaining superior consistency with the global composition.

\begin{table}[t]
\centering
\setlength{\tabcolsep}{3pt}
\begin{tabular}{lcccc}
\hline
Model & PSNR$\uparrow$ & PSNR-M$\uparrow$ & SSIM$\uparrow$ & SSIM-M$\uparrow$  \\
\hline
SketchEdit~\cite{Sketchedit}  & 18.49 & 10.66 & 0.853 & 0.601 \\ 
BrushNet~\cite{Brushnet}      & 16.17 & 7.85  & 0.798 & 0.502 \\ 
MagicQuill~\cite{Magicquill}  & 17.78 & 9.19  & 0.868 & 0.583 \\ 
\textbf{Ours}                 & \textbf{19.03} & \textbf{11.41} & \textbf{0.872}  & \textbf{0.688} \\
\hline
\end{tabular}
\caption{
Additional quantitative results for Line-guided Region Redrawing. 
PSNR-M and SSIM-M denote PSNR and SSIM computed only within the masked region.
}
\label{tab:supp_psnr_ssim}
\end{table}

\subsection{VIEScore Prompt Configuration}
For the automated evaluation of Instruction-Guided Editing, we utilized the VIEScore~\cite{ku2024viescore} protocol with Gemini-2.0-Flash and Qwen3-VL-30B. As dictated by the evaluation protocol, the multimodal model returns a list of two sub-scores for both \textbf{Semantic Consistency (SC)} and \textbf{Perceptual Quality (PQ)}. To compute the final metric for a single image, we take the minimum of its two sub-scores for both SC and PQ, thereby strictly penalizing any singular failure mode (e.g., severe artifacts or significant over-editing). The \textbf{Overall} score per image is then computed as the geometric mean of its final SC and PQ scores ($\sqrt{\text{SC} \times \text{PQ}}$), ensuring that high performance is required across both semantic adherence and visual quality. Finally, the metrics reported in our quantitative tables represent the arithmetic mean of these image-level scores across the entire test set. To ensure reproducibility, we provide the exact system prompts used for this evaluation in Table~\ref{tab:viescore_prompts}.

\begin{table*}[ht]
    \centering
    \small
    \caption{\textbf{System Prompts for VIEScore Evaluation.} The detailed instructions and scoring criteria provided to the MLLM.}
    \label{tab:viescore_prompts}
    \begin{tabular}{l p{0.8\linewidth}}
        \toprule
        \textbf{Metric} & \textbf{System Prompt / Instruction} \\
        \midrule
        \textbf{Perceptual Quality (PQ)} & 
        RULES: \newline
        The image is an AI-generated image. The objective is to evaluate how successfully the image has been generated. \newline \newline
        \textbf{From scale 0 to 10:} \newline
        A score from 0 to 10 will be given based on \textbf{image naturalness}. \newline
        (0 indicates that the scene in the image does not look natural at all or gives an unnatural feeling such as wrong sense of distance, or wrong shadow, or wrong lighting. 10 indicates that the image looks natural.) \newline \newline
        A second score from 0 to 10 will rate the \textbf{image artifacts}. \newline
        (0 indicates that the image contains a large portion of distortion, or watermark, or scratches, or blurred faces, or unusual body parts, or subjects not harmonized. 10 indicates the image has no artifacts.) \newline \newline
        Put the score in a list such that output score = [naturalness, artifacts] \\
        \midrule
        \textbf{Semantic Consistency (SC)} & 
        \textbf{From scale 0 to 10:} \newline
        A score from 0 to 10 will be given based on the \textbf{success of the editing}. \newline
        (0 indicates that the scene in the edited image does not follow the editing instruction at all. 10 indicates that the scene in the edited image follows the editing instruction text perfectly.) \newline \newline
        A second score from 0 to 10 will rate the \textbf{degree of overediting} in the second image. \newline
        (0 indicates that the scene in the edited image is completely different from the original. 10 indicates that the edited image can be recognized as a minimally edited yet effective version of original.) \newline \newline
        Put the score in a list such that output score = [score1, score2], where `score1' evaluates the editing success and `score2' evaluates the degree of overediting. \newline \newline
        Editing instruction: \texttt{<instruction>} \\
        \bottomrule
    \end{tabular}
\end{table*}

\section{Comparison of Data Generation Methods}

To justify the necessity of our proposed data generation pipeline—which integrates generation with strict post-filtering—we conducted a study to evaluate the quality of the constructed training pairs.
We compared the editing pairs produced by our pipeline against those generated by standard paradigms. To ensure a rigorous comparison, we employed state-of-the-art generative backbones (SDXL~\cite{podell2023sdxl} and FLUX) for these baseline methods.
Our goal is to demonstrate that our pipeline provides significantly more accurate and spatially consistent training data, establishing a superior foundation for training the SketchAssist model compared to:

\begin{enumerate}
    \item \textbf{Prompt-to-Prompt Editing}~\cite{hertz2022prompt, brooks2023instructpix2pix}:  
    This method manipulates image content by modifying the text prompt while keeping the model's cross-attention maps aligned with the original image (implemented via SDXL). By adding, removing, or replacing certain words in the prompt, local semantic changes can be introduced without altering unrelated image regions. 
    
    \item \textbf{Object detection based Local Editing} (adapted from~\cite{yu2025anyedit}):  
    This approach first detects objects or body parts in the image using \emph{Grounding Dino}~\cite{liu2024grounding} and \emph{SAM2}~\cite{ravi2024sam}. The detected region is then erased and refilled via an inpainting model.  To represent the state-of-the-art, we evaluated two variants of this paradigm: one using \emph{SDXL-Inpainting} and another using \emph{FLUX.1-Fill}.
\end{enumerate}

\paragraph{Prompt-to-Prompt:}

\begin{figure*}[!ht]
\centering
\includegraphics[width=1.0\linewidth]{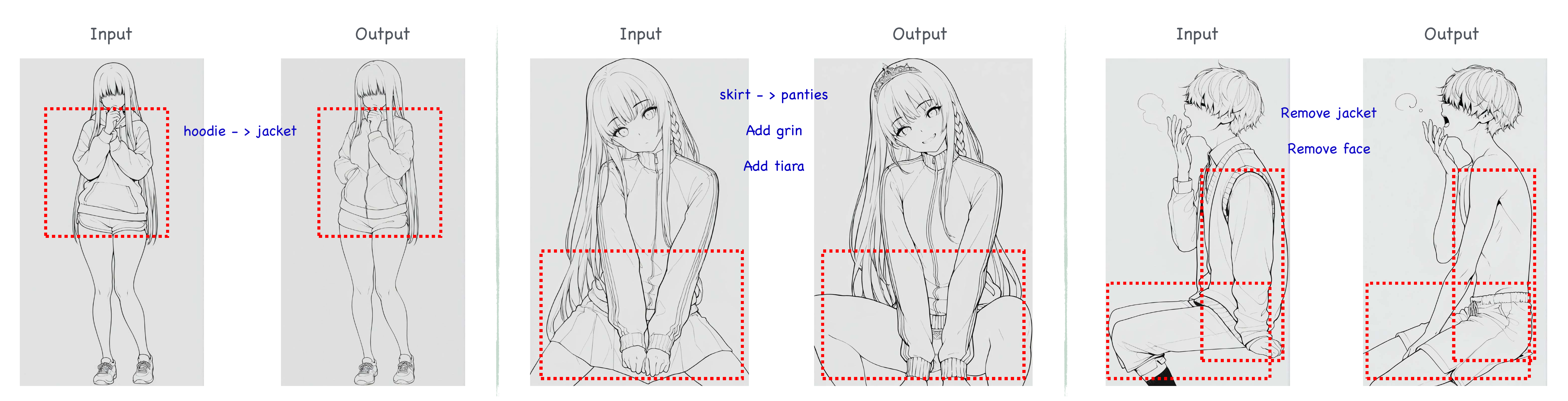}
\caption{Failures of generated edit pairs using Prompt-to-Prompt~\cite{hertz2022prompt, brooks2023instructpix2pix}. 
While Prompt-to-Prompt supports compositional edits by adding, replacing, or removing attributes in the prompt, it often fails to preserve consistent human poses and scene composition across edits. 
The examples illustrate typical cases where pose or layout consistency is lost despite correct attribute modifications. 
Regions enclosed by red dashed boxes indicate areas where the human pose has changed between the source and edited image.}
\label{fig:failure_p2p}
\end{figure*}

We constructed a total of 101 editing pairs using the Prompt-to-Prompt method~\cite{hertz2022prompt}, following the data generation paradigm widely utilized by InstructPix2Pix~\cite{brooks2023instructpix2pix}. 
Editing instructions were randomly sampled with lengths between 1 and 5 operations, enabled by Prompt-to-Prompt’s ability to sequentially modify multiple words or phrases within a single prompt—effectively simulating multi-step edits in one generation. 
Participants rated each pair as success or failure.  
As shown in Table~\ref{tab:p2p_success}, our method achieves a substantially higher average success rate (87.88\%) compared to Prompt-to-Prompt (46.89\%) under the same evaluation protocol.
Fig.~\ref{fig:failure_p2p} illustrates typical failure cases of Prompt-to-Prompt. 
While Prompt-to-Prompt supports compositional edits by adding, replacing, or removing attributes in the prompt, it often fails to maintain consistent human poses and scene composition across edits. 
In the examples shown, although the target attributes are correctly modified, the resulting images exhibit noticeable changes in body posture or spatial arrangement, breaking the intended continuity between the original and edited versions.

\begin{table}[h]
\centering
\begin{tabular}{cc}
\toprule
\textbf{Method} & \textbf{Average Success Rate (\%)} \\
\midrule
P2P   & 46.89 \\
Ours  & 87.88 \\
\bottomrule
\end{tabular}
\caption{User study success rate (\%) for Prompt-to-Prompt editing pairs compared with our method.}
\label{tab:p2p_success}
\end{table}

\paragraph{Object Detection + Inpainting:}
\begin{figure*}[!ht]
\centering
\includegraphics[width=1.0\linewidth]{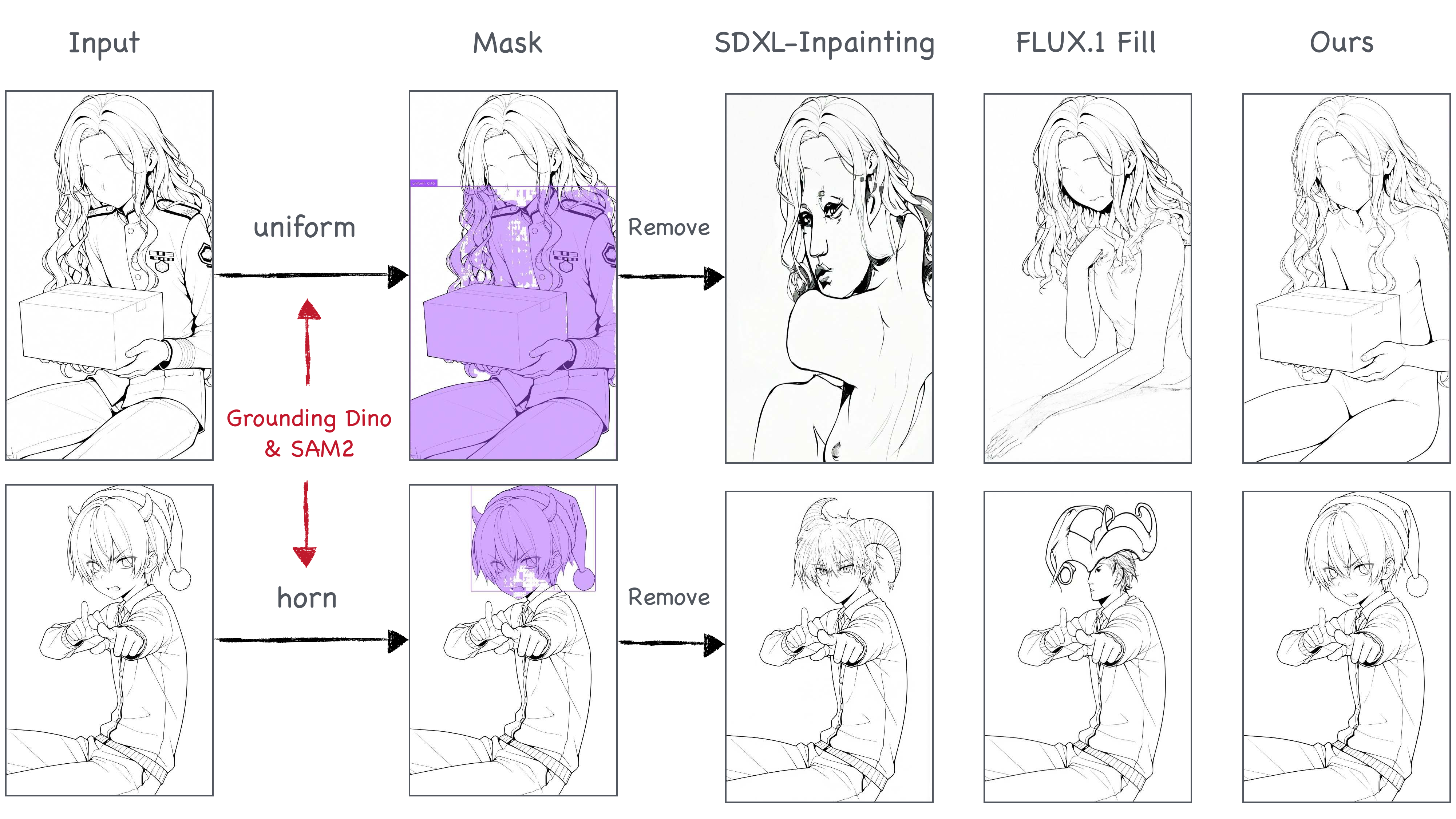}
\caption{Qualitative comparisons between our method and an object-detection-based inpainting approach (SDXL-Inpainting and FLUX.1-Fill), where both baselines generate content based on segmentation masks, while our method synthesizes results via an addition sequence. In this setting, the detected target region is used to construct a “remove” edit pair by erasing the region and filling it via inpainting. Two common failure cases of the object-detection-based pipeline are illustrated: 
Top — when the segmented region is overly large (e.g., covering clothing), the inpainting process fails to recover the original human pose; 
Bottom — under coarse sketch guidance, Grounding Dino and SAM2 produces inaccurate segmentation masks, leading to incorrect inpainting regions.}

\label{fig:failure_seg}
\end{figure*}
We constructed a total of 64 editing items using an object-detection-based local editing approach adapted from~\cite{yu2025anyedit}. \emph{Grounding Dino}~\cite{liu2024grounding} and \emph{SAM2}~\cite{ravi2024sam} were used to detect target regions, which were then erased or replaced via inpainting. Although the detection-based pipeline can be extended to multi-step edits by repeatedly detecting and erasing different regions, such repetition increases complexity and may introduce structural inconsistencies. Therefore, we restrict our comparison to the single-step case.

The quantitative results in Table~\ref{tab:det_success} show that our method achieves a significantly higher average success rate (91.50\%) compared to SDXL-inpainting (16.28\%) and FLUX.1-fill (35.31\%).

Beyond the quantitative advantage, Fig.~\ref{fig:failure_seg} presents qualitative comparisons between our method and an object-detection-based approach. In this setting, the detected target region is used to construct a “remove” edit pair by erasing the region and filling it via inpainting. Two typical failure cases of object-detection-based inpainting are shown: (1) when the segmented region is too large (e.g., covering clothing), the inpainting process fails to restore the original human pose; and (2) when guided by a coarse sketch, SAM2 produces inaccurate segmentation masks, resulting in incorrect inpainting regions.

\begin{table}[h]
\centering
\begin{tabular}{lc}
\toprule
\textbf{Method} & \textbf{Average Success Rate (\%)} \\
\midrule
SDXL-inpainting & 16.28 \\
FLUX.1-fill     & 35.31 \\
Ours            & 91.50 \\
\bottomrule
\end{tabular}
\caption{User study average success rate (\%) for Object Detection + Inpainting editing pairs, compared with our method.}
\label{tab:det_success}

\end{table}

\section{More Details of Data Generation}
\label{sec:more_dataset}

\subsection{Instruction-guided Editing Data}
\paragraph{Attribute Addition Sequence:} We generate $\mathbf{10{,}000}$ distinct base sketches using anime-style \emph{Stable Diffusion XL (SDXL)} models. Each base sketch is then extended into multiple distinct atomic addition sequences. Each sequence begins with the initial sketch state and is expanded through multiple successive attribute-adding chains. At each step of the sequence, the sketch generated in the preceding step is fed into a \emph{lineart-conditioned ControlNet~\cite{Zhang_2023_ICCV}} as structural guidance, thereby preserving the underlying line structure while allowing semantic content to evolve according to the specified attributes.

\paragraph{Filtering Process:} We apply the \emph{HumanArt}~\cite{ju2023human} model in conjunction with CLIP-based~\cite{Clip} similarity scoring to evaluate structural changes between source--target pairs. Pairs exhibiting significant composition, pose, or structural changes are discarded. Furthermore, we utilize the WD14 Tagger~\cite{wd14tagger} to confirm the precise presence or absence of target attributes, and employ Qwen-VL to validate the overall alignment with the editing instruction. Only pairs that pass both the structural-consistency check and the attribute-correctness verification are retained, resulting in a curated subset of roughly $\mathbf{100{,}000}$ images.

\paragraph{Style Diversification:} To expand stylistic coverage, we trained an attribute-removal model based on FLUX.1-Kontext~\cite{FLUX} using our synthetic line art sequences and applied it to approximately 4,000 diverse real-world sketches. 
Crucially, the remover generalizes effectively to these unseen styles by leveraging the backbone's in-context generation design, where texture synthesis is driven by surrounding pixel cues rather than solely on fine-tuned weights.
This stands in contrast to the \emph{addition} task, which requires hallucinating new content and is prone to overfitting to the synthetic line art domain of the source data. 
Thus, these generated multi-style pairs act as essential \textbf{style regularizers}, preventing domain overfitting and enabling SketchAssist to support diverse artistic renditions.

\paragraph{Simulated Guidance Lines:} We first extract line art structure from a diverse collection of source sketches using the anime lineart preprocessor provided in ControlNet~\cite{Zhang_2023_ICCV}, capturing the essential contours of the subject.
Based on these structural maps, we employ \emph{Stable Diffusion XL (SDXL)} combined with a \emph{line art-conditioned ControlNet} to synthesize the final guidance images. To emulate the imperfect quality of real hand-drawn hints, we incorporate prompt modifiers such as ``bad quality'' and ``sketch'',  intentionally producing output lines with lower visual fidelity and jittery strokes.
Crucially, this synthetic data construction introduces a domain gap between the rough guidance lines and the precise geometry of the original sketch. This forces the model to learn semantic correspondence rather than rigid pixel alignment, enabling it to robustly correct user stroke errors and spatial misalignments during inference.

\paragraph{Mask Generation:} As described in the main text, body-part masks are detected using \emph{HumanArt}~\cite{ju2023human}, and object masks are obtained using \emph{Grounding Dino} and \emph{SAM2}.

\section{Quantitative Verification of Input Encoding and Mask Fidelity}
\label{sec:vae_verification}

Our framework employs a unified input strategy where the sketch ($R$), mask ($G$), and guidance ($B$) are encoded via the repurposed RGB channels of the pre-trained VAE. While this design significantly streamlines the architecture, the fidelity of the \textbf{mask encoding} is paramount. In the context of local editing, the binary mask acts as the definitive hard constraint that delimits the editing scope. Therefore, preserving the precise spatial geometry of the mask during VAE encoding/decoding is the absolute prerequisite for successful instruction adherence and structural preservation.

To verify that our input strategy meets this critical requirement, we conducted a rigorous quantitative analysis on the mask regions of our entire test set ($N=200$). Specifically, we passed the composite inputs through the VAE, applied a standard binary threshold ($\tau=0.5$) to the reconstructed $G$-channel, and computed the \textbf{Intersection-over-Union (IoU)} against the ground-truth masks.

\textbf{Results.} As shown in Table~\ref{tab:vae_iou}, the model achieves a near-perfect \textbf{Mean IoU of 0.9995}. Even in the worst-case scenario, the \textbf{Minimum IoU} remains at \textbf{0.9982}. These results confirm that despite the domain gap of the pre-trained VAE, the spatial definition of the editable region is preserved with effectively zero loss. This ensures that the DiT backbone receives accurate localization guidance, establishing a solid foundation for precise local redrawing.

\begin{table}[h]
    \centering
    \vspace{2mm}
    \begin{tabular}{|l|c|c|c|}
        \hline
        \textbf{Metric} & \textbf{Mean} & \textbf{Min} & \textbf{Std} \\
        \hline
        IoU & \textbf{0.9995} & 0.9982 & 0.0003 \\
        \hline
    \end{tabular}
    \caption{\textbf{Quantitative analysis of Mask Reconstruction Quality.} Evaluated on the test set ($N=200$). The near-perfect IoU scores demonstrate that the proposed encoding strategy incurs negligible loss in spatial localization accuracy, ensuring precise editing control.}
    \label{tab:vae_iou}
\end{table}
 
\section{Training Configuration}
\paragraph{Cross-Sequence Sampling}
During training, we employ a cross-sequence sampling strategy, in which edit pairs are constructed by randomly composing between \(\mathbf{1\text{--}5}\) atomic operations from a predefined set (e.g., \emph{add}, \emph{remove}, \emph{replace}). Accordingly, the resulting edit pairs span an edit distance range of \(D = 1\) to \(5\), as defined in the main paper. This design enables the model to learn from both simple single-operation edits and more complex multi-operation transformations, while remaining compatible with the unified framework.

\paragraph{Task-guided Mixture-of-Experts (T-MoE) Architecture}
The \emph{Shared LoRA} module is configured with a fixed rank of $24$, and four \emph{expert LoRA} modules are instantiated with a rank of $12$ each. The router selects the \emph{top-2} experts for each instruction based on a learned gating mechanism conditioned on concatenated text and visual features. This design allows the model to distinguish between text-based editing mode and line-based local redrawing mode, while sharing low-level style and structure representations. For fairness in the ablation study, the non-MoE baseline is configured with a single LoRA module of rank $48$, such that the total number of activated LoRA parameters matches that of the T-MoE configuration.

\section{More Qualitative Comparison}
\label{sec:more_qua}

\begin{figure*}[!ht]
  \centering
   \includegraphics[width=1.0\linewidth]{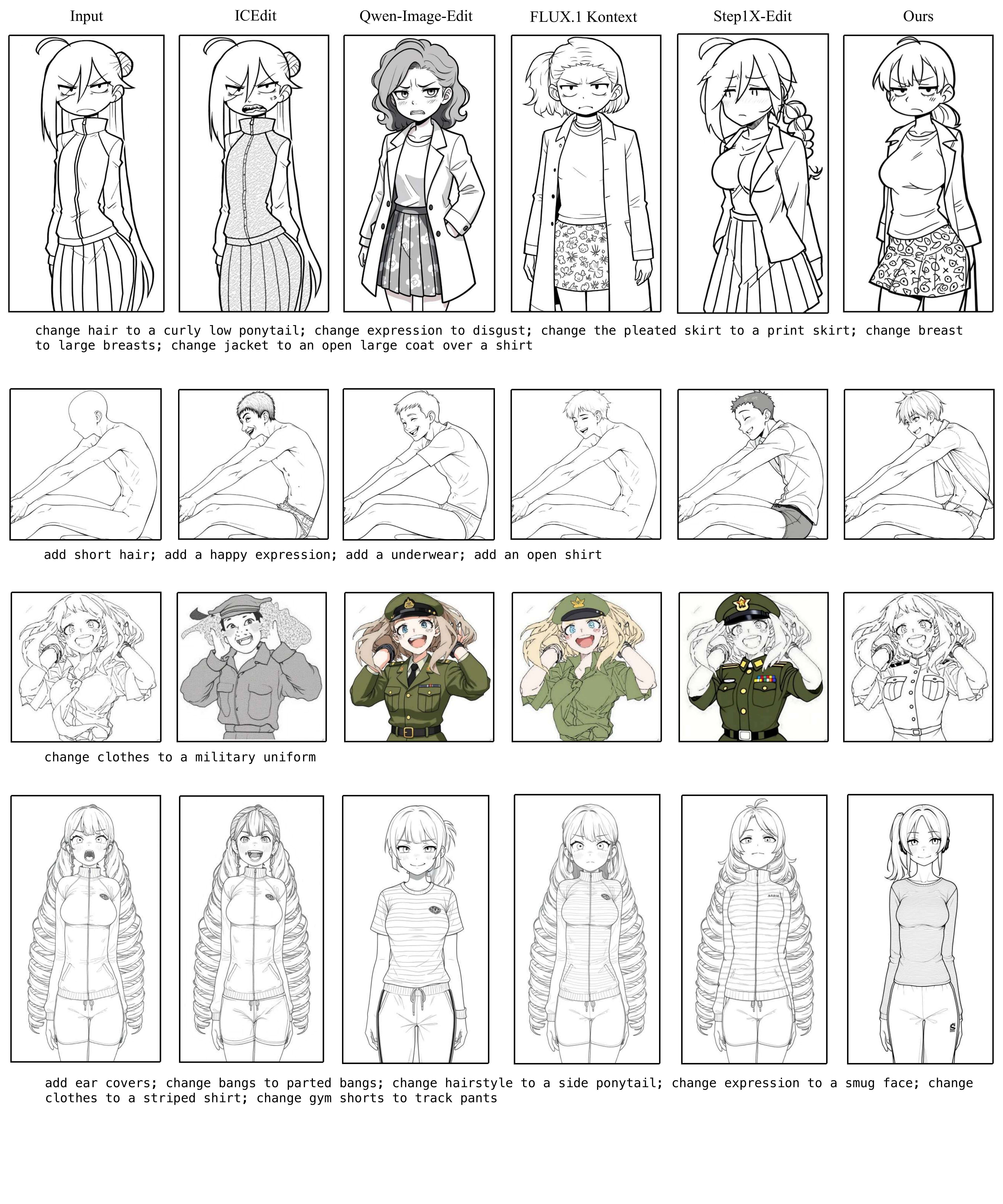}
   \caption{Qualitative comparisons of instruction-based editing. The text beneath each example corresponds to the editing instruction applied.}
   \label{fig:qua_comp_supp1}
\end{figure*}

\begin{figure*}[!ht]
  \centering
   \includegraphics[width=1.0\linewidth]{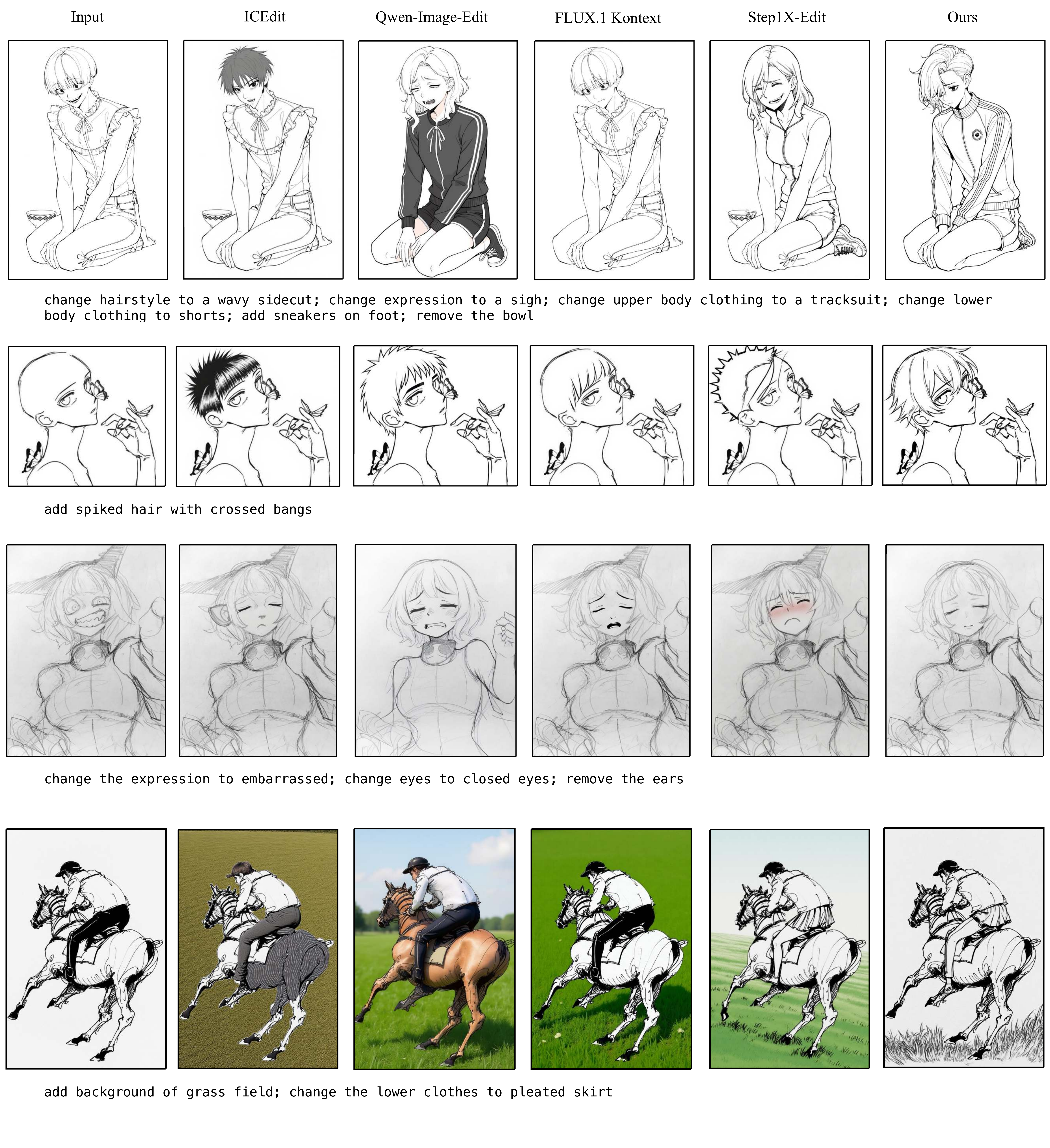}
   \caption{Qualitative comparisons of instruction-based editing. The text beneath each example corresponds to the editing instruction applied.}
    \label{fig:qua_comp_supp2}
\end{figure*}

We provide more qualitative comparisons of instruction-guided editing task with ICEdit~\cite{Enabling}, Qwen-Image-Edit~\cite{Qwen}, FLUX.1 Kontext~\cite{FLUX}, and Step1X-Edit~\cite{liu2025step1x} in Fig.~\ref{fig:qua_comp_supp1} and Fig.~\ref{fig:qua_comp_supp2}. In these experiments, we deliberately adopt complex and multi-faceted editing instructions, requiring simultaneous modifications to multiple attributes, objects, or regions. Our method demonstrates a strong ability to accurately follow these instructions while preserving the original scene composition and human pose, without introducing unintended structural changes. This highlights both the advantage of our curated data construction and the efficacy of our Cross-Sequence Sampling strategy in handling multi-instruction edits within a single generation.

\begin{figure*}[!ht]
  \centering
   \includegraphics[width=1.0\linewidth]{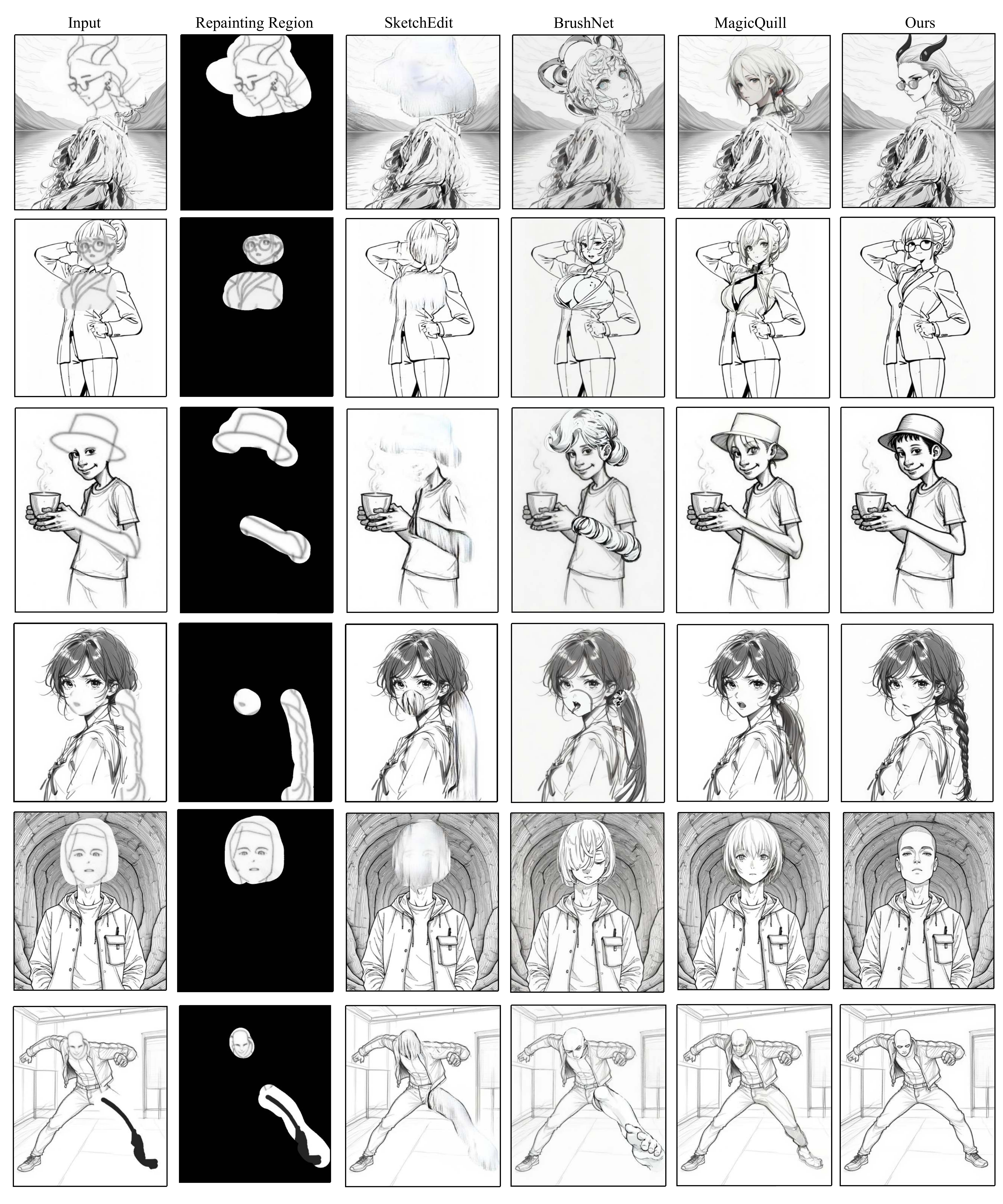}
   \caption{Qualitative comparisons of Line-guided Redrawing}
   \label{fig:qua_comp_supp3}
\end{figure*}

We further provide qualitative comparisons on a line-guided region redrawing task—where the target mask is specified and structural guidance is provided via line drawings—with SketchEdit~\cite{Sketchedit}, BrushNet~\cite{Brushnet}, and MagicQuill~\cite{Magicquill} in Fig.~\ref{fig:qua_comp_supp3}. SketchEdit is designed for local sketch-to-image translation and therefore cannot operate directly on complete sketch images. BrushNet and MagicQuill can perform local redrawing on sketch images, but in our evaluation they fail to reliably follow the provided guidance lines, producing structures misaligned with the intended line geometry. Our method, by comparison, redraws the masked region in strict accordance with the guidance lines while maintaining high stylistic fidelity to the original image, ensuring structural accuracy and visual coherence.

\end{document}